\documentclass[conference]{IEEEtran}
\usepackage[T1]{fontenc}
\usepackage[utf8]{inputenc}
\usepackage{cite}
\usepackage{amsmath,amssymb,amsfonts}
\usepackage{booktabs}
\usepackage{multirow}
\usepackage{graphicx}
\usepackage{xcolor}
\usepackage{array}
\usepackage{url}
\usepackage{hyperref}
\usepackage{balance}

\usepackage{algorithm}
\usepackage{algpseudocode}

\def\BibTeX{{\rm B\kern-.05em{\sc i\kern-.025em b}\kern-.08em
    T\kern-.1667em\lower.7ex\hbox{E}\kern-.125emX}}

\usepackage{blindtext}
\usepackage{eso-pic}
\usepackage{eso-pic}

\makeatletter

\def\ps@IEEEtitlepagestyle{%
  \def\@oddfoot{\mycopyrightnotice}%
  \def\@evenfoot{}%
}
\def\mycopyrightnotice{%
  {\parbox{\textwidth}{\scriptsize \copyright~2026 IEEE. Personal use of this material is permitted. Permission from IEEE must be obtained for all other uses, in any current or future media, including reprinting/republishing this material for advertising or promotional purposes, creating new collective works, for resale or redistribution to servers or lists, or reuse of any copyrighted component of this work in other works.}}%
  \gdef\mycopyrightnotice{}
}

\hypersetup{hidelinks}
\title{\vspace*{1cm}Dual-Adaptive Z-Score PPO: Dynamic Clipping Using Action Confidence and Normalized Advantage Magnitude}
\author{
\IEEEauthorblockN{Taha M. Mahmoud, Naima Kaabouch}
\IEEEauthorblockA{Artificial Intelligence Research (AIR) Center\\
University of North Dakota\\
Grand Forks, ND, USA\\
Email: taha.mahmoud@und.edu, naima.kaabouch@und.edu}
}
\begin{document}
\maketitle

\begin{abstract}
Proximal Policy Optimization (PPO) remains a popular on-policy reinforcement learning algorithm owing to its stability and ease of implementation. Its clipping threshold, however, is fixed, so every training sample receives the same trust-region width regardless of how reliable the corresponding policy update is. This work examines adaptive-clipping variants of PPO in which the constant clipping parameter is replaced by sample-dependent thresholds computed from policy and return statistics.

The central method, Dual-Adaptive Z-Score PPO (PPO-DAZ3), sets the clipping interval for each sample from two signals: action confidence, given by the probability of the selected action under the current policy, and update confidence, given by the normalized advantage magnitude. The rest of the standard PPO training pipeline is left unchanged.

Evaluation on three classic control benchmarks (LunarLander-v3, CartPole-v1, and Acrobot-v1) indicates that adaptive clipping improves learning in some settings but not all. PPO-DAZ3 performs best on LunarLander-v3, the smooth advantage-based variant PPO-AZ3S leads on Acrobot-v1, and standard PPO stays competitive on CartPole-v1, suggesting that adaptive clipping is most beneficial on harder control tasks.

\end{abstract}


\section{Introduction}
Among policy-gradient methods, Proximal Policy Optimization (PPO) has become a standard choice in modern reinforcement learning, as it approximates trust-region learning in a practical way while keeping implementation simple \cite{schulman2017ppo,schulman2015trpo}. The clipped form of PPO bounds the policy ratio $r_t(\theta)$ with a fixed threshold $\epsilon$; this guards against destructive updates and permits several epochs of minibatch optimization on a single rollout batch \cite{schulman2017ppo}. PPO is commonly paired with generalized advantage estimation (GAE), which lowers gradient variance and stabilizes learning \cite{schulman2016gae}.

Standard PPO, however, relies on a static clipping threshold. All samples in a rollout share one trust-region width, even though some updates rest on strong evidence and others on noisy or uncertain estimates. This observation motivates a simple question: does PPO benefit from a clipping interval that adapts at the level of individual samples? Adaptive or learned trust-region mechanisms have been studied before, for example PPO-$\lambda$ \cite{chen2018ppolambda}, PPO-RPE \cite{kobayashi2023pporpe}, and SARD-PPO \cite{zhao2023sardppo}. These works demonstrate that adaptive control of clipping can be advantageous, yet the space of possible designs is far from exhausted.

Most existing adaptive PPO variants steer the clipping interval with a single signal, such as state importance, density-ratio estimates, or sample reuse. The formulation studied here instead uses two signals, policy confidence and advantage magnitude, that jointly set the clipping interval. Under this design, the trust region widens only when both the policy-side and return-side signals point to a reliable update, and narrows when either one indicates uncertainty.

Reinforcement learning is being applied in a growing range of domains, including robotics, control systems, and cybersecurity, and recent work has also combined advanced machine learning methods with security analytics and intrusion detection \cite{mahmoudLightweightFederatedSupervised2025,mahmoudQuantumSecureVotingSystem2025}.

The study centres on a small, controlled family of adaptive-clipping variants that share a common PPO backbone. The first variant, \textbf{PPO-AZ3}, groups samples into three $z$-score buckets according to normalized advantage magnitude and assigns each bucket a \texttt{low}, \texttt{mid}, or \texttt{high} clipping threshold. The second, \textbf{PPO-AZ3S}, replaces this three-step rule with a smooth sigmoid mapping, removing the discontinuities at bucket boundaries. The principal contribution, \textbf{PPO-DAZ3} (Dual-Adaptive Z-Score PPO), extends the idea by fusing two signals:

\begin{enumerate}
    \item \textbf{Action confidence}, obtained from the current policy's probability of the chosen action.
    \item \textbf{Advantage confidence}, obtained from the normalized absolute advantage magnitude.
\end{enumerate}

The underlying intuition is simple: a sample that combines high action confidence with strong advantage evidence can afford a somewhat wider trust region, whereas samples with low confidence or weak evidence should stay tightly clipped.

The three variants form a deliberate, progressive design study. PPO-AZ3 serves as a simple advantage-based adaptive-clipping baseline, PPO-AZ3S adds a smooth formulation that avoids discontinuities, and PPO-DAZ3 adds a policy-side confidence signal on top. Building the methods in this order makes it possible to isolate the effect of each design element.

The scope of the paper is deliberately empirical rather than theoretical. Every algorithm uses the same architecture, optimizer, rollout size, and training schedule, so differences in outcome can be attributed to the clipping rule alone. Such control matters because implementation details have a large influence on PPO behaviour and are easily mistaken for algorithmic improvements. DAZ3 is therefore presented conservatively, as a reproducible adaptive-clipping mechanism supported by encouraging initial evidence, not as a general-purpose replacement for PPO.

The contributions of this work are as follows:

\begin{itemize}

\item A progressive family of adaptive-clipping PPO variants driven by normalized advantage statistics, comprising a bucket-based formulation (AZ3) and a smooth continuous formulation (AZ3S) for adapting the clipping threshold per sample.

\item \textbf{PPO-DAZ3}, a dual-signal adaptive-clipping mechanism that combines action confidence and advantage magnitude into sample-specific clipping thresholds without altering the standard PPO objective.

\item A controlled empirical comparison on three benchmark environments (LunarLander-v3, CartPole-v1, and Acrobot-v1) with an identical PPO backbone, isolating the effect of adaptive clipping on learning performance.

\end{itemize}

\noindent\textbf{Paper Organization:} Section \ref{sec:sec2} surveys related work on PPO and adaptive clipping. Section \ref{sec:sec3} details the methodology, covering the AZ3, AZ3S, and DAZ3 variants. Section~IV describes the experimental setup, including implementation details and the evaluation protocol. Section \ref{sec:sec4} presents and discusses the results, Section \ref{sec:sec5} covers limitations and directions for future research, and Section \ref{sec:sec6} concludes the paper.

\section{Related Work}\label{sec:sec2}

PPO was introduced as a first-order alternative to trust-region policy optimization (TRPO), aiming to retain the stability of constrained policy updates while avoiding second-order optimization and expensive line searches \cite{schulman2017ppo,schulman2015trpo}. GAE is commonly paired with PPO to improve the quality of policy gradients by controlling the bias--variance trade-off in the advantage estimator \cite{schulman2016gae}.

A growing body of work has explored \emph{adaptive} trust-region or clipping strategies for PPO. PPO-$\lambda$ introduced an adaptive clipping mechanism motivated by state importance and policy improvement targets \cite{chen2018ppolambda}. PPO-RPE replaced the raw density ratio with a symmetric relative Pearson formulation and derived the clipping threshold from the estimated error scale of the policy update \cite{kobayashi2023pporpe}. More recently, SARD-PPO proposed a combination of sample-adaptive reuse and dual clipping to improve stability in robotic control tasks \cite{zhao2023sardppo}. These studies collectively suggest that a fixed clipping threshold may not always be optimal.

Table~\ref{tab:adaptive_clipping_comparison} summarizes key differences between representative adaptive clipping PPO variants and the method proposed in this work. Prior approaches vary primarily in the signals used to adjust the clipping threshold and in the level at which the adaptation is applied (update-level, batch-level, or sample-level).

\begin{table}[t]
\centering
\caption{Comparison of adaptive clipping strategies in PPO variants}
\label{tab:adaptive_clipping_comparison}
\renewcommand{\arraystretch}{1.1}
\setlength{\tabcolsep}{4pt}
\begin{tabular}{p{2.0cm} p{2.3cm} p{1.0cm} p{2.5cm}}
\toprule
\textbf{Method} & \textbf{Signal} & \textbf{Level} & \textbf{Mechanism} \\
\midrule
PPO-$\lambda$ \cite{chen2018ppolambda} 
& State importance 
& Update 
& Adaptive clipping \\

PPO-RPE \cite{kobayashi2023pporpe} 
& Density-ratio error 
& Batch 
& Pearson-ratio threshold \\

SARD-PPO \cite{zhao2023sardppo} 
& Sample reuse 
& Sample 
& Dual clipping \\

\textbf{DAZ3 (ours)} 
& Action conf. + adv. mag. 
& Sample 
& Dual-signal clipping \\
\bottomrule
\end{tabular}
\end{table}

The proposed approach differs from prior adaptive clipping methods in two key aspects. First, clipping thresholds are determined at the \emph{sample level} using normalized advantage statistics rather than global or batch-level update metrics. Second, the DAZ3 formulation integrates two independent signals: (i) \emph{policy-side confidence}, measured from the probability of the selected action under the current policy, and (ii) \emph{return-side evidence}, captured through normalized advantage magnitude. This dual-signal formulation allows the clipping interval to adapt simultaneously to policy certainty and value-based learning signals while preserving the standard PPO objective.

Finally, the reinforcement learning reproducibility literature highlights that apparent improvements can be highly sensitive to implementation details, environment stochasticity, and evaluation protocols \cite{henderson2017deep}. For this reason, the present study emphasizes controlled comparisons under an identical PPO backbone and reports results as preliminary multi-seed evidence rather than definitive state-of-the-art claims.

\section{Methodology}\label{sec:sec3}
\subsection{PPO Baseline}
For a state--action pair $(s_t,a_t)$, PPO optimizes the clipped surrogate objective
\begin{equation}
L^{\text{clip}}(\theta)=
\mathbb{E}_t
\left[
\min\left(
r_t(\theta)\hat{A}_t,
\text{clip}(r_t(\theta),1-\epsilon,1+\epsilon)\hat{A}_t
\right)
\right],
\end{equation}
where
\begin{equation}
r_t(\theta)=\frac{\pi_\theta(a_t|s_t)}{\pi_{\theta_{\text{old}}}(a_t|s_t)}.
\end{equation}
In standard PPO, $\epsilon$ is constant over all samples \cite{schulman2017ppo}. In this study, all variants use the same actor--critic network, optimizer, rollout size, minibatch size, epoch count, entropy coefficient, value-loss coefficient, and gradient clipping. Only the rule for determining the sample clipping threshold changes.

The present work explores a progressive family of adaptive clipping mechanisms designed to investigate how different sources of information can guide sample-wise clipping adjustments. The first variant, AZ3, introduces a simple bucket-based adaptation rule based on normalized advantage statistics. The second variant, AZ3S, replaces the discrete bucket structure with a smooth sigmoid mapping to reduce discontinuities between samples. The final method, DAZ3, extends this formulation by incorporating an additional policy-side signal derived from action confidence.

\subsection{PPO-AZ3}
AA-PPO-AZ3 computes normalized absolute advantages
\begin{equation}
z_t = \frac{|\hat{A}_t|-\mu_{|\hat{A}|}}{\sigma_{|\hat{A}|}+\delta},
\end{equation}
where $\mu_{|\hat{A}|}$ and $\sigma_{|\hat{A}|}$ are the batch mean and standard deviation of $|\hat{A}_t|$, and $\delta$ is a small numerical constant. Samples are then partitioned into three buckets:
\begin{equation}
\epsilon_t=
\begin{cases}
\epsilon_{\text{low}}, & z_t < z_0,\\
\epsilon_{\text{mid}}, & z_0 \le z_t < z_1,\\
\epsilon_{\text{high}}, & z_t \ge z_1.
\end{cases}
\end{equation}
This mechanism preserves PPO's clipped objective but allows update aggressiveness to vary across samples.

\subsection{PPO-AZ3S}
AZ3S keeps the same intuition but removes hard bucket discontinuities. The normalized advantage score is passed through a sigmoid function,
\begin{equation}
s_t = \sigma\left(k(z_t-c)\right),
\end{equation}
and then linearly mapped into the clipping interval:
\begin{equation}
\epsilon_t = \epsilon_{\text{low}} + 
(\epsilon_{\text{high}}-\epsilon_{\text{low}}) s_t.
\end{equation}
This produces a smooth adaptive threshold, reducing Sudden sample-to-sample changes.

\subsection{PPO-DAZ3}
The proposed DAZ3 extends AZ3S by using two confidence signals.

\textbf{Action Confidence:}
Let $p_t=\pi_\theta(a_t|s_t)$ be the probability of the chosen action under the current policy, and let $u=1/|\mathcal{A}|$ be the uniform baseline for a discrete action space. DAZ3 computes an action-confidence score
\begin{equation}
S_t^{\text{act}} = \sigma\left(k_a(p_t-u)\right).
\end{equation}
This score increases when the selected action is strongly preferred over a uniform guess.

\textbf{Advantage Confidence:}
Using the same normalized advantage statistic $z_t$, DAZ3 computes
\begin{equation}
S_t^{\text{adv}} = \sigma\left(k_v(z_t-c_v)\right).
\end{equation}
This score is high when the absolute advantage is large relative to the batch distribution.

\textbf{Combined Dual Score: }
The two signals are fused as
\begin{equation}
S_t^{\text{dual}} = \alpha S_t^{\text{act}} + (1-\alpha) S_t^{\text{adv}},
\end{equation}
where $\alpha \in [0,1]$ controls the balance between policy-side and return-side evidence. The final sample-specific clipping threshold is
\begin{equation}
\epsilon_t = \epsilon_{\text{low}} +
(\epsilon_{\text{high}}-\epsilon_{\text{low}})S_t^{\text{dual}}.
\end{equation}
The PPO objective is unchanged except that the static $\epsilon$ is replaced by $\epsilon_t$ inside the clipping operator.

The PPO objective remains unchanged except that the static clipping parameter $\epsilon$ is replaced by the sample-specific threshold $\epsilon_t$ inside the clipping operator.

Algorithm~\ref{alg:daz3} summarizes the DAZ3 adaptive clipping procedure used during each PPO update.

\begin{algorithm}[t]
\caption{DAZ3 Dual-Adaptive Clipping}
\label{alg:daz3}
\begin{algorithmic}[1]
\Require Batch $\{(s_t,a_t,\hat{A}_t)\}_{t=1}^{N}$, policy $\pi_\theta$, action space size $|\mathcal{A}|$, clipping bounds $\epsilon_{\text{low}}, \epsilon_{\text{high}}$, sigmoid parameters $k_a,k_v,c_v$, mixing weight $\alpha$, small constant $\delta$

\State Compute absolute advantages $|\hat{A}_t|$
\State Compute batch statistics $\mu_{|\hat{A}|}$ and $\sigma_{|\hat{A}|}$
\State Set uniform baseline $u = 1/|\mathcal{A}|$

\For{$t = 1$ to $N$}

\State $z_t \gets \dfrac{|\hat{A}_t| - \mu_{|\hat{A}|}}{\sigma_{|\hat{A}|}+\delta}$

\State $p_t \gets \pi_\theta(a_t|s_t)$

\State $S_t^{act} \gets \sigma\!\left(k_a(p_t-u)\right)$

\State $S_t^{adv} \gets \sigma\!\left(k_v(z_t-c_v)\right)$

\State $S_t^{dual} \gets \alpha S_t^{act} + (1-\alpha)S_t^{adv}$

\State $\epsilon_t \gets \epsilon_{\text{low}} + (\epsilon_{\text{high}}-\epsilon_{\text{low}})S_t^{dual}$

\EndFor

\State Use $\epsilon_t$ in PPO clipped objective

\end{algorithmic}
\end{algorithm}

\subsection{Experimental Setup}
\textbf{Implementation: }
All experiments were implemented in PyTorch using Gymnasium environments \cite{towers2024gymnasium}. The policy and value function share a two-layer multilayer perceptron with hidden dimension 128 and \texttt{tanh} activations. Training used Adam with learning rate $3\times10^{-4}$, GAE with $\gamma=0.99$ and $\lambda=0.95$, rollout size 1024, 4 training epochs per batch, minibatch size 64, entropy coefficient 0.01, value coefficient 0.5, and gradient norm clipping at 0.5. The default PPO clipping threshold was 0.20. For the adaptive methods, the low/mid/high bounds were 0.195, 0.200, and 0.205.

\textbf{Benchmarks: }
Three discrete-action benchmarks were used:
\begin{itemize}
    \item \textbf{LunarLander-v3}: a sparse and moderately difficult control task;
    \item \textbf{CartPole-v1}: a simple benchmark that saturates quickly;
    \item \textbf{Acrobot-v1}: a harder swing-up task with negative episodic returns.
\end{itemize}
CartPole-v1 was trained for 100k timesteps; LunarLander-v3 and Acrobot-v1 were trained for 200k timesteps. Each algorithm was run with seeds 42, 43, and 44.

\textbf{Evaluation: }
A deterministic greedy policy was evaluated every 10,240 environment steps using 10 episodes. Final results were recorded using an independent evaluation seed schedule. Because only three seeds were available in the current draft, all findings are reported as preliminary. This is consistent with the recommendation that deep RL claims should be made carefully when variance is high \cite{henderson2017deep}.

\section{Results and Discussion}\label{sec:sec4}

\subsection{Aggregate Multi-Seed Performance}

Table~\ref{tab:main_results} summarizes the final evaluation return averaged across three random seeds. Higher values indicate better performance, except for Acrobot-v1 where less negative returns correspond to better control.

\begin{table*}[t]
\centering
\caption{Final return across three seeds. Mean and standard deviation are computed from the final evaluation return of each seed. Higher is better; for Acrobot-v1, less negative is better.}
\label{tab:main_results}
\begin{tabular}{lcccc}
\toprule
Environment & PPO & PPO-AZ3 & PPO-AZ3S & PPO-DAZ3 \\
\midrule
LunarLander-v3 & 19.57 $\pm$ 72.59 & 90.96 $\pm$ 103.76 & 42.92 $\pm$ 21.22 & \textbf{123.33 $\pm$ 26.26} \\
CartPole-v1    & \textbf{387.77 $\pm$ 121.68} & 256.03 $\pm$ 52.07 & 319.77 $\pm$ 156.70 & 185.13 $\pm$ 17.35 \\
Acrobot-v1     & -90.67 $\pm$ 5.21 & -83.13 $\pm$ 4.93 & \textbf{-77.07 $\pm$ 3.83} & -83.67 $\pm$ 12.38 \\
\bottomrule
\end{tabular}
\end{table*}

Table~\ref{tab:main_results} shows that the best-performing method depends on the environment. PPO-DAZ3 achieves the strongest mean return on LunarLander-v3, PPO remains strongest on CartPole-v1, and PPO-AZ3S performs best on Acrobot-v1. These environment-specific outcomes suggest that adaptive clipping is beneficial in some settings but not universally superior to standard PPO.

To better understand these outcomes, Figures~\ref{fig:acrobot_curve}--\ref{fig:lunarlander_curve} illustrate the learning dynamics during training. In each case, the solid lines represent the mean evaluation return across seeds and the shaded regions indicate cross-seed variability. Figure~\ref{fig:final_performance} complements these learning curves by summarizing the final evaluation performance across all environments using mean return and standard deviation.

\subsection{Learning Dynamics Across Environments}

\textbf{Acrobot-v1: }

\begin{figure}[t]
\centering
\includegraphics[width=0.9\linewidth]{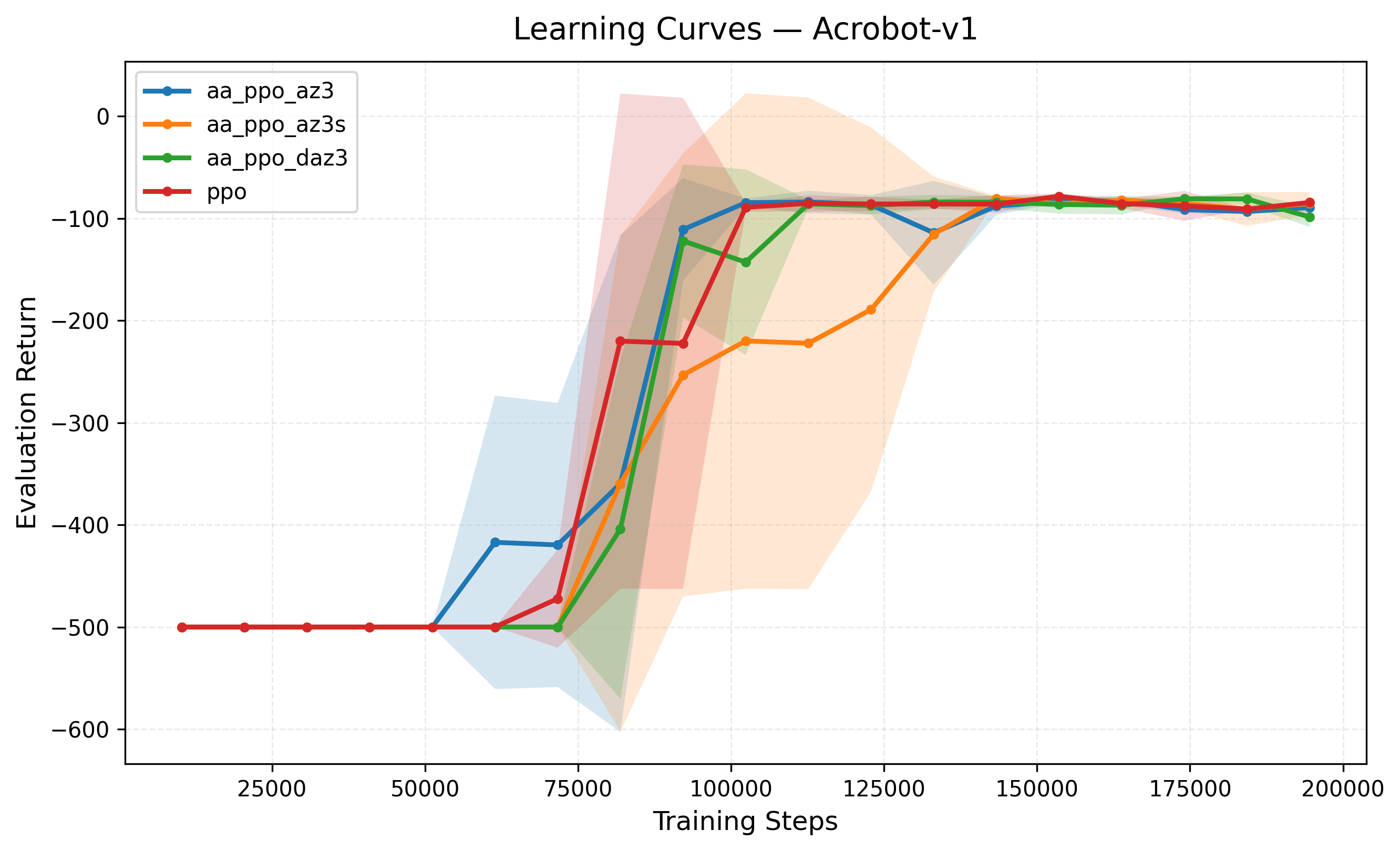}
\caption{Learning curves for Acrobot-v1. Shaded regions indicate variability across seeds. PPO-AZ3S achieves the strongest final performance on this task.}
\label{fig:acrobot_curve}
\end{figure}

Figure~\ref{fig:acrobot_curve} shows the training curves for Acrobot-v1. All methods begin near the minimum return of approximately $-500$, indicating that the agent initially fails to perform the swing-up task. Around 80k--100k training steps, all algorithms begin improving as the policy starts discovering effective momentum-building strategies.

\begin{figure*}[t]
\centering
\includegraphics[width=\textwidth]{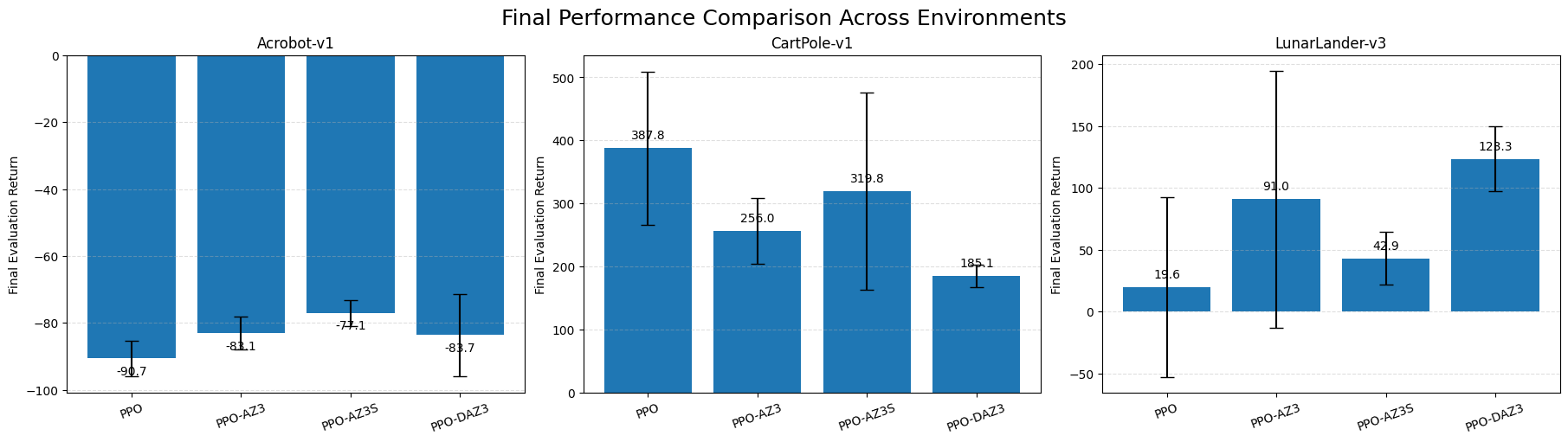}
\caption{Final evaluation performance across the three benchmark environments. Bars represent the mean return averaged across three seeds, and error bars indicate the corresponding standard deviation. PPO-AZ3S achieves the strongest performance on Acrobot-v1, PPO remains strongest on CartPole-v1, and PPO-DAZ3 produces the highest mean return on LunarLander-v3.}
\label{fig:final_performance}
\end{figure*}

Among the adaptive methods, PPO-AZ3S demonstrates the strongest final performance and one of the most stable convergence patterns. This observation is consistent with Table~\ref{tab:main_results} and the Acrobot-v1 panel in Figure~\ref{fig:final_performance}, where PPO-AZ3S achieves the best mean return ($-77.07$). The smooth advantage-based adaptation appears beneficial for this task because it avoids sudden threshold changes that may arise in the bucket-based AZ3 method. In contrast, PPO-DAZ3 performs similarly to PPO-AZ3 but does not provide additional improvement in this environment, suggesting that the action-confidence signal does not add substantial value for this particular control problem.

\textbf{CartPole-v1: }

\begin{figure}[t]
\centering
\includegraphics[width=0.9\linewidth]{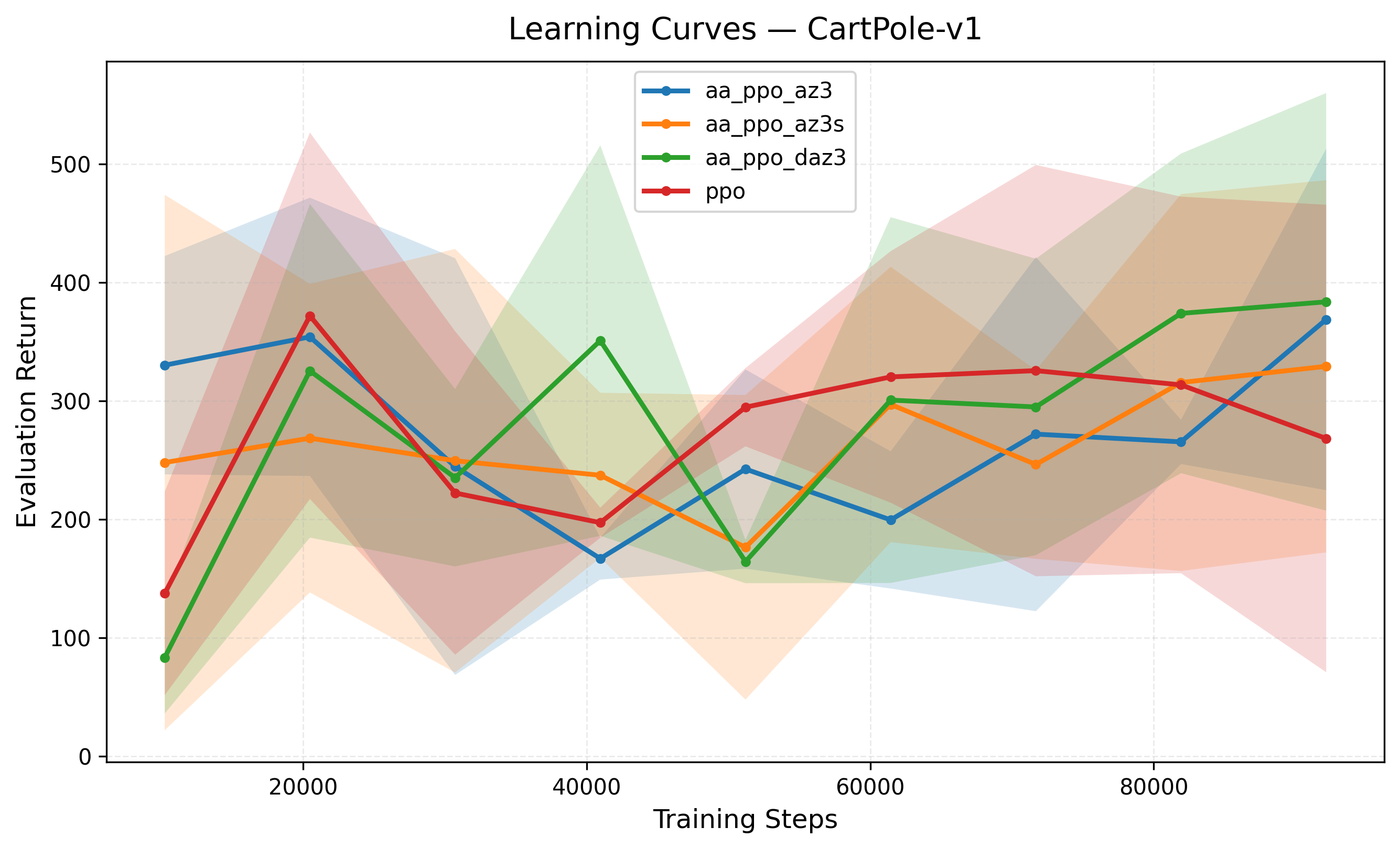}
\caption{Learning curves for CartPole-v1. Standard PPO remains highly competitive on this saturated benchmark.}
\label{fig:cartpole_curve}
\end{figure}

Figure~\ref{fig:cartpole_curve} presents the results for CartPole-v1, a relatively simple control benchmark that is often solved quickly by policy-gradient methods. Standard PPO reaches strong performance early in training and maintains the highest final average return.

Adaptive clipping methods do not provide clear improvements on this task and in some cases introduce additional variance during training. This behavior is expected because CartPole-v1 rapidly approaches near-optimal performance even with a fixed clipping threshold. Consequently, dynamically adjusting the trust-region width offers limited benefit and may instead introduce unnecessary fluctuations in policy updates. This interpretation is also supported by Figure~\ref{fig:final_performance}, where PPO shows the highest final mean return among the four methods.

\textbf{LunarLander-v3: }

\begin{figure}[t]
\centering
\includegraphics[width=0.9\linewidth]{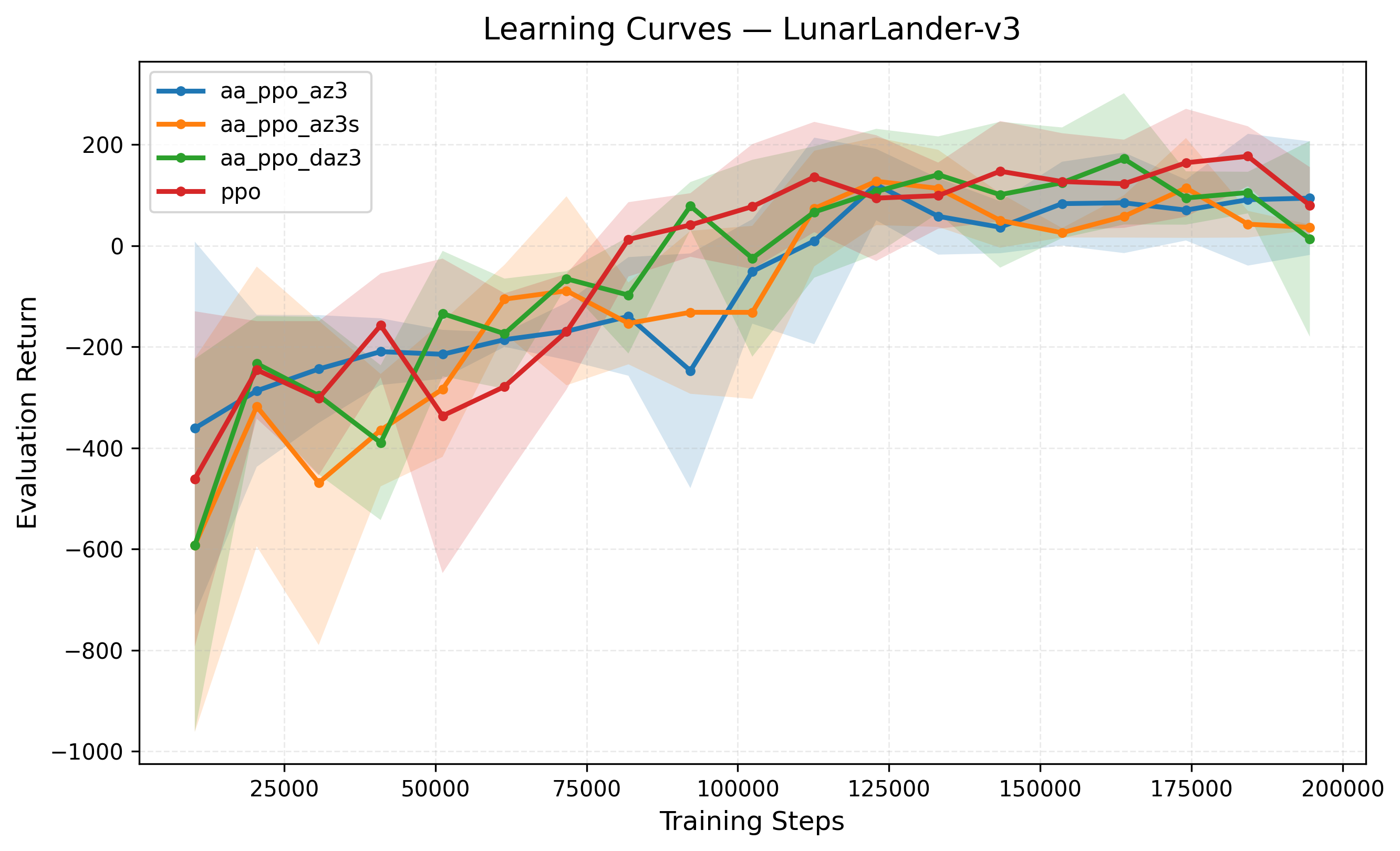}
\caption{Learning curves for LunarLander-v3. PPO-DAZ3 demonstrates stronger improvement in the later stages of training.}
\label{fig:lunarlander_curve}
\end{figure}

Figure~\ref{fig:lunarlander_curve} illustrates the learning behavior on LunarLander-v3, which is a more complex and sparse-reward environment. All methods start with highly negative returns, reflecting the difficulty of the initial exploration phase. As training progresses, the adaptive variants begin improving faster than the PPO baseline.

The proposed PPO-DAZ3 method eventually achieves the highest mean performance, reaching an average return of $123.33$ compared with $19.57$ for PPO. The improvement becomes most apparent after approximately 100k training steps, where PPO-DAZ3 maintains stronger late-stage performance while exhibiting lower cross-seed variation than PPO-AZ3. This same trend is visible in Figure~\ref{fig:final_performance}, where PPO-DAZ3 has the highest final bar on LunarLander-v3.

These results suggest that combining action-confidence and advantage-magnitude signals can provide a useful mechanism for adapting the clipping threshold in more challenging environments. When both signals indicate reliable learning evidence, the algorithm can safely allow larger policy updates, improving learning without destabilizing training.

\subsection{Interpretation}

Taken together, the results reveal several consistent patterns.

\textbf{Environment difficulty matters.} Adaptive clipping appears most beneficial in more challenging environments such as LunarLander-v3, where different samples may require different trust-region widths.

\textbf{Smooth adaptation can improve stability.} PPO-AZ3S performs particularly well on Acrobot-v1, supporting the idea that smooth mappings from advantage statistics to clipping thresholds can reduce sudden policy changes.

\textbf{Dual-signal adaptation is environment dependent.} PPO-DAZ3 performs best on LunarLander-v3, where policy confidence and advantage magnitude together provide meaningful signals for controlling update size. However, this additional signal does not consistently translate into improvements on simpler tasks such as CartPole-v1.

\textbf{Final ranking alone is insufficient.} The learning curves show that some methods may exhibit similar final returns while following different learning trajectories. For this reason, both training dynamics and final performance are necessary to interpret the behavior of adaptive clipping variants.

These results indicate that adaptive clipping is environment-dependent, providing the greatest benefit in more challenging tasks where the optimal trust-region width varies across samples.

\subsection{Novelty and Positioning}

The contribution of this work is not the general idea of adaptive clipping for PPO, which has already been explored in several forms \cite{chen2018ppolambda,kobayashi2023pporpe,zhao2023sardppo}. Instead, the novelty lies in the formulation of a \emph{dual-signal} adaptive clipping rule that combines policy-side information (action confidence) with return-side information (normalized advantage magnitude) while preserving the standard PPO training pipeline.

Unlike prior adaptive PPO variants that rely on single-signal or global adaptation strategies, the proposed method introduces a dual-signal formulation that combines policy-side and return-side information at the sample level.

Because the method does not alter the PPO objective, backbone architecture, optimizer, or rollout pipeline, it can be integrated easily into existing PPO implementations. This simplicity makes the approach easy to reproduce, easy to ablate, and suitable for follow-up evaluation on larger benchmark suites and continuous-control environments.

\section{Discussion, Limitations, and Future Work}\label{sec:sec5}

While the proposed dual-adaptive clipping mechanism demonstrates promising results, several limitations should be considered when interpreting the findings.

First, the experimental scale remains limited. The current evaluation reports results averaged over three random seeds. Although sufficient to illustrate consistent trends, a more comprehensive study should include additional seeds, confidence intervals, and statistical significance tests. Additional metrics such as learning-curve area under the curve (AUC), sample efficiency, and wall-clock training time would provide a more complete understanding of algorithm behavior. In this study, the analysis focuses on final evaluation returns due to availability.

Second, the benchmark suite is restricted to classic discrete-action control environments. While CartPole-v1, Acrobot-v1, and LunarLander-v3 are widely used, they represent relatively small-scale problems compared with modern continuous-control benchmarks. In particular, CartPole-v1 is a saturated task where standard PPO often achieves near-optimal performance, limiting the observable benefit of adaptive clipping. These results indicate that adaptive clipping is environment-dependent and most beneficial in more challenging tasks where the optimal trust-region width varies across samples.

Third, the present work does not include a detailed ablation study of the DAZ3 hyperparameters, including the weighting coefficient $\alpha$ and scaling factors of the dual signals. Systematic ablation experiments would help clarify the contribution of each component.

Finally, the current formulation of the action-confidence signal is designed for discrete-action policies. Extending the method to continuous-action settings is an important direction for future work. Adapting the confidence signal for Gaussian policies would enable evaluation on MuJoCo benchmarks such as Hopper, Walker2d, and HalfCheetah. Despite these limitations, the results provide consistent empirical evidence that combining action-confidence and advantage-magnitude signals can improve PPO clipping adaptation in certain environments.
\section{Conclusion} \label{sec:sec6}

This paper introduced a family of adaptive clipping strategies for Proximal Policy Optimization (PPO) and proposed AA-PPO-DAZ3, a dual-signal adaptive clipping mechanism that combines action-confidence and advantage-magnitude signals to dynamically adjust the clipping threshold during training. The approach was evaluated on three classic reinforcement learning benchmarks while maintaining a fixed PPO backbone to ensure fair comparison. Experimental results indicate that adaptive clipping can improve learning in certain environments. In particular, PPO-DAZ3 achieved the strongest performance on LunarLander-v3, while the smooth advantage-based variant PPO-AZ3S performed best on Acrobot-v1. On CartPole-v1, standard PPO remained competitive, suggesting that adaptive clipping is not universally beneficial and may be most useful in environments where trust-region flexibility is needed. Future work will expand the empirical evaluation, including more seeds, statistical testing, broader benchmarks, and extensions to continuous-action reinforcement learning tasks.
\bibliographystyle{IEEEtran}
\bibliography{references}

\begin{thebibliography}{10}
\providecommand{\url}[1]{#1}
\csname url@samestyle\endcsname
\providecommand{\newblock}{\relax}
\providecommand{\bibinfo}[2]{#2}
\providecommand{\BIBentrySTDinterwordspacing}{\spaceskip=0pt\relax}
\providecommand{\BIBentryALTinterwordstretchfactor}{4}
\providecommand{\BIBentryALTinterwordspacing}{\spaceskip=\fontdimen2\font plus
\BIBentryALTinterwordstretchfactor\fontdimen3\font minus
  \fontdimen4\font\relax}
\providecommand{\BIBforeignlanguage}[2]{{%
\expandafter\ifx\csname l@#1\endcsname\relax
\typeout{** WARNING: IEEEtran.bst: No hyphenation pattern has been}%
\typeout{** loaded for the language `#1'. Using the pattern for}%
\typeout{** the default language instead.}%
\else
\language=\csname l@#1\endcsname
\fi
#2}}
\providecommand{\BIBdecl}{\relax}
\BIBdecl

\bibitem{schulman2017ppo}
\BIBentryALTinterwordspacing
J.~Schulman, F.~Wolski, P.~Dhariwal, A.~Radford, and O.~Klimov, ``Proximal
  {Policy} {Optimization} {Algorithms},'' Aug. 2017, arXiv:1707.06347 [cs].
  [Online]. Available: \url{http://arxiv.org/abs/1707.06347}
\BIBentrySTDinterwordspacing

\bibitem{schulman2015trpo}
\BIBentryALTinterwordspacing
J.~Schulman, S.~Levine, P.~Abbeel, M.~Jordan, and P.~Moritz,
  ``\BIBforeignlanguage{en}{Trust {Region} {Policy} {Optimization}},'' in
  \emph{\BIBforeignlanguage{en}{Proceedings of the 32nd {International}
  {Conference} on {Machine} {Learning}}}.\hskip 1em plus 0.5em minus
  0.4em\relax PMLR, Jun. 2015, pp. 1889--1897. [Online]. Available:
  \url{https://proceedings.mlr.press/v37/schulman15.html}
\BIBentrySTDinterwordspacing

\bibitem{schulman2016gae}
\BIBentryALTinterwordspacing
J.~Schulman, P.~Moritz, S.~Levine, M.~Jordan, and P.~Abbeel,
  ``High-{Dimensional} {Continuous} {Control} {Using} {Generalized} {Advantage}
  {Estimation},'' Oct. 2018, arXiv:1506.02438 [cs]. [Online]. Available:
  \url{http://arxiv.org/abs/1506.02438}
\BIBentrySTDinterwordspacing

\bibitem{chen2018ppolambda}
\BIBentryALTinterwordspacing
G.~Chen, Y.~Peng, and M.~Zhang, ``An {Adaptive} {Clipping} {Approach} for
  {Proximal} {Policy} {Optimization},'' Apr. 2018, arXiv:1804.06461 [cs].
  [Online]. Available: \url{http://arxiv.org/abs/1804.06461}
\BIBentrySTDinterwordspacing

\bibitem{kobayashi2023pporpe}
\BIBentryALTinterwordspacing
T.~Kobayashi, ``Proximal policy optimization with adaptive threshold for
  symmetric relative density ratio,'' \emph{Results in Control and
  Optimization}, vol.~10, p. 100192, Mar. 2023. [Online]. Available:
  \url{https://www.sciencedirect.com/science/article/pii/S2666720722000649}
\BIBentrySTDinterwordspacing

\bibitem{zhao2023sardppo}
\BIBentryALTinterwordspacing
L.-y. Zhao, T.-q. Chang, J.~Zhang, L.~Zhang, K.-x. Chu, L.-b. Guo, and D.-p.
  Kong, ``A policy optimization algorithm based on sample adaptive reuse and
  dual-clipping for robotic action control,'' \emph{Applied Soft Computing},
  vol. 134, p. 109967, Feb. 2023. [Online]. Available:
  \url{https://www.sciencedirect.com/science/article/pii/S156849462201016X}
\BIBentrySTDinterwordspacing

\bibitem{mahmoudLightweightFederatedSupervised2025}
\BIBentryALTinterwordspacing
T.~M. Mahmoud and N.~Kaabouch, ``Lightweight {Federated} {Supervised} {Machine}
  {Learning} for {IoT} {Botnet} {Detection},'' in \emph{2025 {International}
  {Conference} on {Artificial} {Intelligence}, {Computer}, {Data} {Sciences}
  and {Applications} ({ACDSA})}, Aug. 2025, pp. 1--6. [Online]. Available:
  \url{https://ieeexplore.ieee.org/abstract/document/11165820}
\BIBentrySTDinterwordspacing

\bibitem{mahmoudQuantumSecureVotingSystem2025}
\BIBentryALTinterwordspacing
------, ``Quantum-{Secure} {Voting} {System}: {Leveraging} {QKD}, {Dual}-{Key}
  {Symmetric} {Encryption}, and {Receipt} {Verification} for {Practical}
  {Privacy} and {Integrity},'' in \emph{2025 {International} {Conference} on
  {Artificial} {Intelligence}, {Computer}, {Data} {Sciences} and {Applications}
  ({ACDSA})}, Aug. 2025, pp. 1--6. [Online]. Available:
  \url{https://ieeexplore.ieee.org/abstract/document/11165862}
\BIBentrySTDinterwordspacing

\bibitem{henderson2017deep}
\BIBentryALTinterwordspacing
P.~Henderson, R.~Islam, P.~Bachman, J.~Pineau, D.~Precup, and D.~Meger,
  ``\BIBforeignlanguage{en}{Deep {Reinforcement} {Learning} {That}
  {Matters}},'' \emph{\BIBforeignlanguage{en}{Proceedings of the AAAI
  Conference on Artificial Intelligence}}, vol.~32, no.~1, Apr. 2018. [Online].
  Available: \url{https://ojs.aaai.org/index.php/AAAI/article/view/11694}
\BIBentrySTDinterwordspacing

\bibitem{towers2024gymnasium}
\BIBentryALTinterwordspacing
M.~Towers, A.~Kwiatkowski, J.~Terry, J.~U. Balis, G.~D. Cola, T.~Deleu,
  M.~Goulão, A.~Kallinteris, M.~Krimmel, A.~KG, R.~Perez-Vicente, A.~Pierré,
  S.~Schulhoff, J.~J. Tai, H.~Tan, and O.~G. Younis, ``Gymnasium: {A}
  {Standard} {Interface} for {Reinforcement} {Learning} {Environments},'' Nov.
  2025, arXiv:2407.17032 [cs]. [Online]. Available:
  \url{http://arxiv.org/abs/2407.17032}
\BIBentrySTDinterwordspacing

\end{thebibliography}

\end{document}